\documentclass[preprint, 5p]{elsarticle}

\usepackage[ruled,vlined]{algorithm2e}
\usepackage{mathbbol}
\usepackage{mathtools}
\usepackage{amssymb, amsmath}
\usepackage{lineno,hyperref}
\usepackage{array}
\usepackage{multirow}
\usepackage{todonotes}
\usepackage{caption}
\usepackage{subcaption}
\usepackage{booktabs}
\modulolinenumbers[5]

\journal{Journal of Web Semantics}

\begin{document}

\newlength\mylen
\newcommand\myinput[1]{%
	\settowidth\mylen{\KwIn{}}%
	\setlength\hangindent{\mylen}%
	\hspace*{0.5cm}#1\\}
\SetKwInOut{Variables}{Variables}
\SetKwRepeat{Do}{do}{while}

\newcommand\mycommfont[1]{\footnotesize\ttfamily{#1}}
\SetCommentSty{mycommfont}

\begin{frontmatter}

\title{Leveraging Wikidata's edit history in knowledge graph refinement tasks}

\author[address]{Alejandro Gonzalez-Hevia\corref{correspondingauthor}}
\cortext[correspondingauthor]{Corresponding author}
\ead[url]{www.alejgh.com}
\ead{uo251513@uniovi.es}

\author[address]{Daniel Gayo-Avello}
\ead[url]{www.danigayo.info}
\ead{dani@uniovi.es}

\address[address]{Department of Computer Science, University of Oviedo, Spain}

\begin{abstract}
Knowledge graphs have been adopted in many diverse fields for a variety of purposes. Most of those applications rely on valid and complete data to deliver their results, pressing the need to improve the quality of knowledge graphs. A number of solutions have been proposed to that end, ranging from rule-based approaches to the use of probabilistic methods, but there is an element that has not been considered yet: the edit history of the graph. In the case of collaborative knowledge graphs (e.g., Wikidata), those edits represent the process in which the community reaches some kind of fuzzy and distributed consensus over the information that best represents each entity, and can hold potentially interesting information to be used by knowledge graph refinement methods. In this paper, we explore the use of edit history information from Wikidata to improve the performance of type prediction methods. To do that, we have first built a JSON dataset containing the edit history of every instance from the 100 most important classes in Wikidata. This edit history information is then explored and analyzed, with a focus on its potential applicability in knowledge graph refinement tasks. Finally, we propose and evaluate two new methods to leverage this edit history information in knowledge graph embedding models for type prediction tasks. Our results show an improvement in one of the proposed methods against current approaches, showing the potential of using edit information in knowledge graph refinement tasks and opening new promising research lines within the field.
\end{abstract}


\end{frontmatter}

\let\linenumbers\nolinenumbers\nolinenumbers

\section{Introduction}
Different fields have incorporated the use of domain-specific knowledge graphs during recent years to solve their tasks. Some concrete examples of such domain-specific tasks include performing investment analysis \cite{ruan2016building}, managing diseases and symptoms from medical records \cite{rotmensch2017learning}, or automatically generating test cases for software projects \cite{nayak2020knowledge}, among many others. Furthermore, the emergence of several open and general-purpose knowledge graphs, such as DBPedia \cite{bizer2009dbpedia} and Wikidata \cite{vrandevcic2012wikidata}, has also attracted new communities closer to the Semantic Web by allowing them to exploit this structured information for many different applications. It goes without saying that most of those applications rely on the correctness and completeness of the data in the knowledge graph to deliver their results.

It is therefore crucial to ensure a high level of quality for those knowledge graphs. This has led to works that define quality metrics and dimensions to better analyze and understand data quality \cite{piscopo2019we}. Those works reveal the existence of constraint violations and missing information in modern knowledge graphs, among other quality problems \cite{farber2018linked, shenoy2022study}. 

Therefore, a number of different approaches have been proposed to improve the quality of knowledge graphs. Some of them follow a deductive approach, where a set of rules or constraints that each triple must follow are defined to enforce data quality \cite{patel2015using, prud2014shape}. Other proposals follow an inductive approach, using predictive models or alternative probabilistic methods to try to fill incomplete information or fix errors in the knowledge graph \cite{paulheim2014improving}.

However, in the specific case of collaborative knowledge graphs like Wikidata, there is an element that has not been fully explored yet: its edit history information. One of the main features of Wikidata, telling it apart from other open general-purpose knowledge graphs, is its collaborative approach: anyone can start from scratch editing entities in Wikidata. At the time of this writing, there have been 1,640,933,943 edits made to Wikidata. In those edits, the community has progressively built a consensus --fuzzy and somewhat distributed among the editors-- over the information that best represents each entity within the knowledge graph, while also capturing the natural evolution of those entities across time.

In this paper we explore the possibilities of leveraging edit information to refine the contents of a knowledge graph, laying the foundations for future work in this area. Our main contributions are:
\begin{enumerate}
    \item The creation of a JSON dataset containing the complete edit history of every entity of the 100 most important classes in Wikidata, following Wikidata's data model (Section \ref{sec:extract-edit-hist}).
    \item An analysis of the main editing patterns from contributors, edits made by class, and divisiveness in Wikidata based on the edit information (Section \ref{sec:analysis-edit-hist}). This information is analyzed with a focus on its possible applications to knowledge graph refinement tasks.
    \item The proposal of two approaches to leverage edit history data in type prediction tasks: the use of edits in the negative sampling process of knowledge graph embeddings models, and using the edit information as labeled data fed to a classifier (Section \ref{sec:type-pred-proposals}). We perform an evaluation of these approaches against a set of baselines and analyze the impact of using edit history information in both approaches.
    \item An RDF dataset containing edit history information about Wikidata, following a custom data format where each operation and revision is serialized to the graph. This RDF dataset is also available without edit history information, and can serve as a baseline to measure the impact of using edit history data in knowledge graph refinement models (Section \ref{sec:type-pred-dataset}).
\end{enumerate}

The rest of this paper is structured as follows. In the next section, we go over Wikidata's data model and provide a formal definition of a knowledge graph and its edits. These concepts are needed to better understand the successive aspects of our work. Section \ref{sec:extract-edit-hist} goes over the process of acquiring edit information from Wikidata. This edit information is explored in section \ref{sec:analysis-edit-hist}. In Section \ref{sec:type-pred}, we propose two new methods to leverage edit information to improve existing knowledge graph embedding models, and we evaluate their performance with respect to current approaches. Related work is reviewed in Section \ref{sec:rel-work}. And, finally, in Section \ref{sec:conclusions} we present the conclusions of this work and future research directions.

\section{Background}
\label{sec2}
\subsection{Wikidata data model}
\textit{Entities} are the basic building block of Wikidata's data model. There are two different types of entities: \textit{items} and \textit{properties}. Each entity is given a unique incremental numeric id, with items being prefixed by a `Q' and properties by a `P'.

A \textit{statement} is composed of a property and a value assigned to that property, optionally having $1$ to $n$ qualifiers and $1$ to $n$ references. In the rest of this paper we will use the term \textit{simple statement} to refer to statements that are just composed of a property and a value. A \textit{statement group} is the set of statements that an item has of a given property. Each entity in Wikidata is composed of $0$ to $n$ statement groups.

\textit{Qualifiers} are used to give further information about a given statement (e.g., the point in time when the statement holds true). Each qualifier is also composed of a property and a value assigned to that property. The combination of a property, value, and qualifier is called a \textit{claim}. \textit{References} are also property-value pairs, and they provide the source that validates a statement.

Aliases, descriptions, and labels constitute the \textit{fingerprint} of an entity. These elements are mapped internally to \texttt{skos:altLabel}, \texttt{schema:description}, and \texttt{rdfs:label} URIs in the RDF representation of an entity. The combination of description and label of an entity in a given language must be unique. An entity can have multiple aliases but only a single description and label for a given language.

\textit{Snaks} are the most basic information structure in Wikidata, and provide information about the value of a property. There are three types of snaks: \textit{value}, \textit{somevalue}, and \textit{novalue}. Value snaks indicate that the property has a known value, which is then represented using Wikidata's available datatypes\footnote{More information available at \url{https://www.wikidata.org/wiki/Special:ListDatatypes}}. Somevalue snaks indicate that the property has a value but its value is unknown\footnote{Somevalue snaks are represented with blank nodes in the RDF serialization of the data model.}. Finally, novalue snaks indicate that the property does not have a value. 

Wikidata also introduces three ranks which can be assigned to each statement: \texttt{preferred}, \texttt{deprecated}, and \texttt{normal}. These ranks are generally used to decide which statements must be returned when querying Wikidata, and also to clean up its user interface when exploring an entity. Statements can also have an order within each statement group. Although the order of statements within each rank is not relevant, it can be changed by users.

All these elements are internally serialized in Wikidata to JSON and different RDF serialization formats. It must be noted that in this section we have covered all the elements that are mentioned in the rest of this paper, but the list is not exhaustive. Additional information about Wikidata's data model and its serialization is available online\footnote{\url{https://www.mediawiki.org/wiki/Wikibase/DataModel?tableofcontents=0}}.

\subsection{Revisions}
Wikidata allows any user to edit entities. Therefore, each entity is composed of a \textit{revision history} that holds all the changes made to the entity by Wikidata contributors. Any of the elements described in the previous section can be changed, and a single revision may hold changes to any number and combination of elements of an entity. Throughout this paper we will use the terms \textit{edition} and \textit{revision} interchangeably, following Wikidata's terminology.

A revision also contains additional metadata, including its timestamp, author of the revision, tags, and a description. Tags are usually used to indicate the device from which the revision was made (or the tool that made the edit, if it was an automated process) and also to indicate the cause of the edit\footnote{A list of the most common tags can be accessed at \url{https://www.wikidata.org/wiki/Special:Tags}} (e.g., to revert vandalism).

In the context of this paper we will use the term \textit{operation} to refer to a single modification made to an entity's element in a revision. One revision may be composed of $1$ to $n$ operations. An operation may represent the addition or removal of a single element from the entity. For the sake of simplicity, we will also consider replacements as operations, which are a combination of an addition and removal operation to an entity.

Wikidata allows \textit{restoring} and \textit{undoing} revisions made to an entity. Restoring allows users to undo all the edits made to an entity up to the selected restoration state. Undoing is more versatile, since it undoes from $1$ to $n$ edits selected by the user, which do not need to be consecutive edits like in the restoration process. In none of those cases the revisions being undone are removed from the revision history of the entity. A new revision is made instead, which includes the necessary operations to undo the selected revisions.

Edits can be manually removed by Wikidata administrators under specific circumstances. These include revisions that contain private information, a violation of copyright, or personal attacks of a serious nature. 

\subsection{Formal definitions}
We will now formally introduce the main elements used throughout this paper. Let $\mathcal{I}$, $\mathcal{L}$ and $\mathcal{B}$ be disjoint countably infinite sets of IRIs, literals and blank nodes respectively. A knowledge graph can be formally defined from a static point of view as a set of triples $(s, p, o) \in (\mathcal{I} \cup \mathcal{B}) \times \mathcal{I} \times (\mathcal{I} \cup \mathcal{L} \cup \mathcal{B})$.

From a dynamic point of view, a knowledge graph is built from a sequence of \textit{operations} $Op = \{op_j : 1 \leq j \leq \infty\}$. Each operation $op_j$ is composed of a triple $t = (s, p, o): s \in (\mathcal{I} \cup \mathcal{B}) \times \mathcal{I} \times (\mathcal{I} \cup \mathcal{L} \cup \mathcal{B})$. $Op^+ = \{t_1, t_2, ..., t_n\}$ represents the set of addition operations of the graph, while $Op^- = \{t'_1, t'_2, ..., t'_m\}$ represents the set of removal operations. The set of all operations is therefore defined as $Op = Op^+ \cup Op^-$. A knowledge graph is built out of $n$ operations, with $K_i$ representing the state of the graph after applying all operations up to operation $i$. Applying an addition operation $op^+_{i+1}=(s,p,o)$ to a graph $K_i$ results in graph $K_{i+1} = K_i \cup (s, p, o)$. On the other hand, applying a removal operation $op^-_{i+1}=(s,p,o)$ to a graph $K_i$ results in graph $K_{i+1} = K_i \setminus (s, p, o)$. The final state of a knowledge graph can be obtained by a successive application of all its operations.

\section{Extracting edit history data from Wikidata}
\label{sec:extract-edit-hist}
We now present the approach followed to extract the edit history information from Wikidata to conduct our experiments.

\subsection{Subset selection}
Wikidata was composed of 97,795,169 entities at the time of this writing, with more than 1,640,933,943 revisions in total made by users\footnote{Source: \url{https://www.wikidata.org/wiki/Wikidata:Statistics}}. Given that working with the entire Wikidata revision history could be too computationally expensive to validate our proposal, we extracted a subset to conduct our experiments.

This subset is composed of instances from the \textit{most important} Wikidata classes. To that end, we have computed the ClassRank \cite{fernandez2021approaches} score of every class in Wikidata, choosing the top 100 classes with the highest score. Then, we extracted the edit history information of every entity that is an instance of any of those classes. We preferred choosing the most important classes for our experiments over producing a random sample since --in general-- entities belonging to central classes receive more attention from the community, and are therefore more promising for exploiting their edit history information. 

To run ClassRank we defined the \textit{P31} property (\textit{instance of}) of Wikidata as a \textit{class-pointer}\footnote{The \textit{class-pointer} is used by ClassRank to fetch those entities from Wikidata that are classes.}. The ClassRank score of a class is computed by aggregating the PageRank \cite{page1999pagerank} scores of all its instances. However, since computing the PageRank score of every entity in Wikidata was too computationally expensive, we used a set of pre-computed PageRank scores. These scores were obtained using the Danker\footnote{\url{https://github.com/athalhammer/danker}} tool, which periodically computes the PageRank score of every existing entity in Wikipedia \cite{Thalhammer2016}. The scores are then mapped from Wikipedia pages to their respective Wikidata entities, getting an approximation of their PageRank value\footnote{These dumps can be accessed at \url{https://danker.s3.amazonaws.com/index.html}}.

The 20 most important classes based on their ClassRank score can be seen in table \ref{tbl:classrank-top20}. These results were then filtered manually, since some of those entities could be considered Wikidata classes at the ontological level, but not at a conceptual one. To do so, we have removed those classes that contained the term ``Wikimedia'' in their labels, since they are used to organize Wikimedia content but do not represent classes at a conceptual level.

\begin{table*}
	\caption{Top 20 most important classes based on their ClassRank score}
	\label{tbl:classrank-top20}
	\centering
	\begin{tabular}{l r r}
	\toprule
    	Name & ClassRank score & Number of instances\\
	\midrule
        human &2,167,439 &3,873,812 \\
        Wikimedia category &1,057,559 &2,207,283 \\
        sovereign state &837,883 &203 \\
        taxon &755,681 &1,962,491 \\
        country &746,208 &193 \\
        point in time with respect to recurrent timeframe &635,401 &3,273 \\
        calendar year &499,551 &666 \\
        big city &321,893 &3,238 \\
        human settlement &321,637 &512,417 \\
        Wikimedia disambiguation page &317,631 &1,294,218 \\
        Wikimedia administration category &302,196 &12,146 \\
        Wikimedia list article &287,605 &301,370 \\
        language &284,433 &8,894 \\
        modern language &257,324 &6,875 \\
        city &247,022 &8,650 \\
        academic discipline &204,426 &1,603 \\
        time zone named for a UTC offset &170,254 &72 \\
        metacategory in Wikimedia projects &167,871 &2,545 \\
        republic &165,881 &78 \\
        capital &161,384 &388 \\
        taxonomic rank &160,822 &67 \\
	\bottomrule
	\end{tabular}
\end{table*}

The final subset is composed of 89 classes and 9.3 million instances --around 10\% of the total number of entities in Wikidata. Although this subset is composed of just a 10\% of the entities in Wikidata, its size is around a 35\% of the total size of Wikidata. This can be explained due to the fact that the most important entities have, in general, more content introduced by the community with respect to lesser important classes.

\subsection{Data extraction}
Wikidata periodically releases public dumps of its contents\footnote{Available at \url{https://dumps.wikimedia.org/wikidatawiki/}}. We have selected the \textit{pages-meta-history} dumps to extract the edit history of every entity from our subset, since these dumps are the only ones containing every revision of each entity and not just their final content. This dataset is composed of several XML files containing metadata of every revision made to each entity, and also a JSON blob with the complete content of the entity after each revision. Our final dataset is built from the \textit{pages-meta-history} dumps from 2021-11-01.

\begin{figure}
\centering
\includegraphics[width=0.485\textwidth]{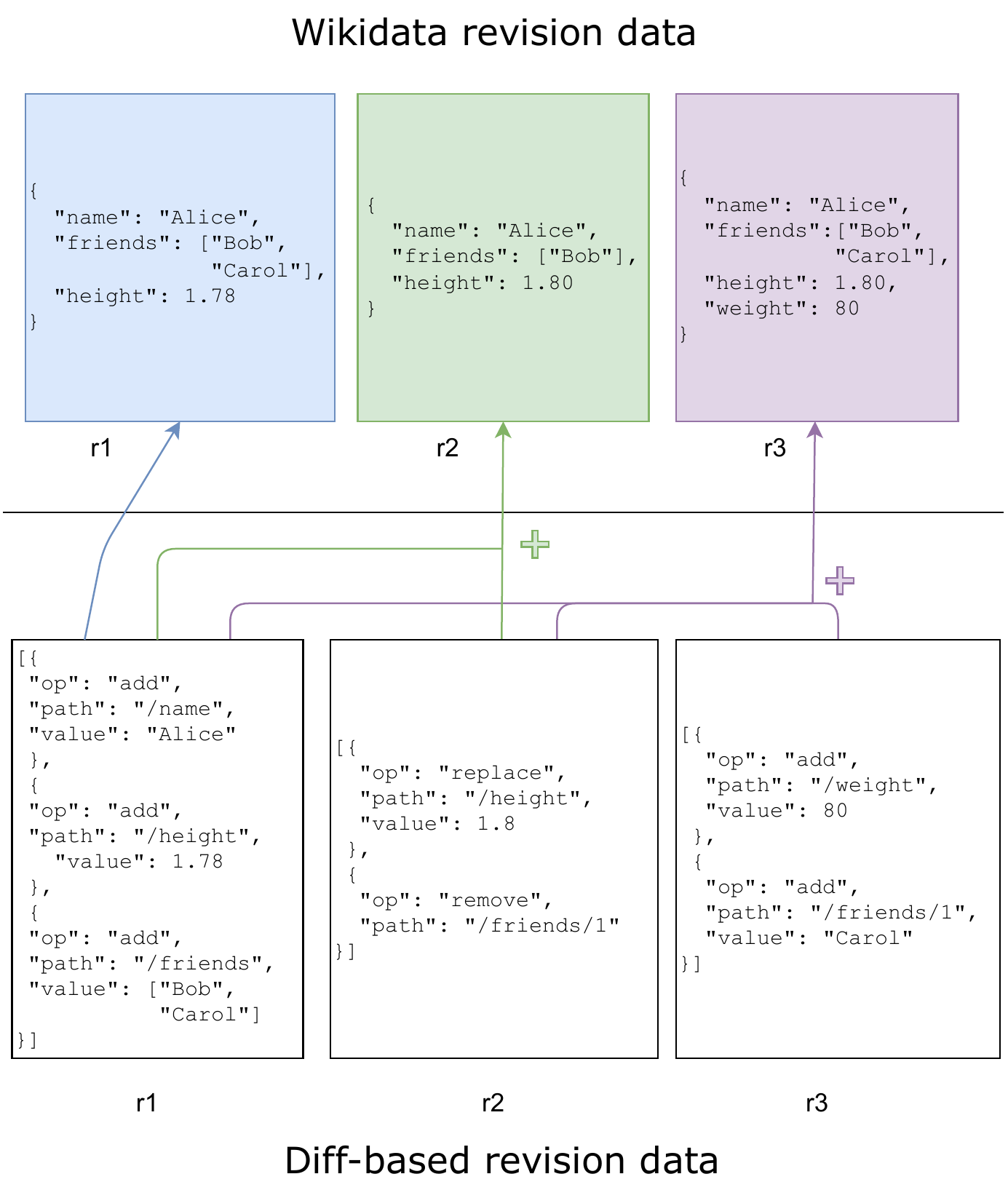}
\caption{Example of revision decompositions into JSON Patch. At the top of the figure we show the simplified content of a Wikidata entity at each revision $r$. At the bottom of the figure we show the decomposition of this entity into diffs in JSON format.}
\label{fig:diffs-decomposition}
\end{figure}

Since working with the complete content of every entity after each revision leads to a lot of redundant entity information, we made some preprocessing steps to reduce the dataset size. Instead of storing the complete JSON content of each entity after every revision $r$, we computed the diff between the JSON content in the previous revision ($r_{t-1}$) and the current one ($r_t$). These diffs are stored in the JSON Patch format\footnote{\url{https://datatracker.ietf.org/doc/html/rfc6902}}, therefore allowing the reconstruction of the entity contents after any revision. To obtain the complete JSON content of an entity at revision $r_t$ we just need to apply the patches of every previous revision up to $r_t$. A simplified example of the decomposition of an entity in diffs is illustrated in figure \ref{fig:diffs-decomposition}.

In this process, the following revision data was extracted:
\begin{itemize}
    \item \textbf{id}: ID of the Wikidata revision.
    \item \textbf{parent\_id}: ID of the previous revision affecting the entity.
    \item \textbf{entity\_id}: ID of the entity altered by the revision.
    \item \textbf{timestamp}: Date of the revision, following the ISO 8601 format (e.g., +2019-05-27T09:31:10Z).
    \item \textbf{username}: Name of the user that made the revision.
    \item \textbf{comment}: Comments of the revision, if they exist.
    \item \textbf{entity\_diff}: Diff of the revision, following the JSON Patch format previously explained.
\end{itemize}

\subsection{Indexing of data}
The revision data previously fetched was then indexed into a \textit{MongoDB} database for further exploration. Table \ref{tbl:edit-dataset-stats} shows the main statistics of the revisions dataset that was indexed.

\begin{table}
	\caption{Statistics of the edits dataset used in the experiments}
	\label{tbl:edit-dataset-stats}
	\centering
	\begin{tabular}{l r}
	\toprule
        \textbf{Dataset size:} & 401GB \\
        \textbf{Number of entities:} & 9,709,099 \\
        \textbf{Number of revisions:} & 443,116,607 \\
        \textbf{Number of operations:} & 986,678,861 \\
	\bottomrule
	\end{tabular}
\end{table}

A custom database architecture was defined in order to optimize the performance of queries needed in the following experiments. The database was split into two main collections: one for \textbf{entities} and another one for \textbf{revisions}. The former contains the final complete JSON content of each entity, while the latter contains the JSON diffs of each entity after each revision. Although the final content of an entity can be obtained by applying all of its revision diffs, this architecture allows for a performant exploration of the revision data and also the current content of each entity.

This dataset has been released in order to promote its exploration by other researchers \cite{gonzalez_hevia_alejandro_2022_6614264}.

\section{Wikidata edits analysis}
\label{sec:analysis-edit-hist}
In this section we explore the edits dataset to exploit the dynamic aspects of Wikidata. While Wikidata is usually explored from a static point of view (i.e., having into account just the state of the knowledge graph at a point in time $t$), having information about its changes lets us understand the community's approach when building the knowledge graph in fine-grained detail. 

\subsection{Life cycle of an entity}
\begin{figure*}
\centering
\includegraphics[width=\textwidth]{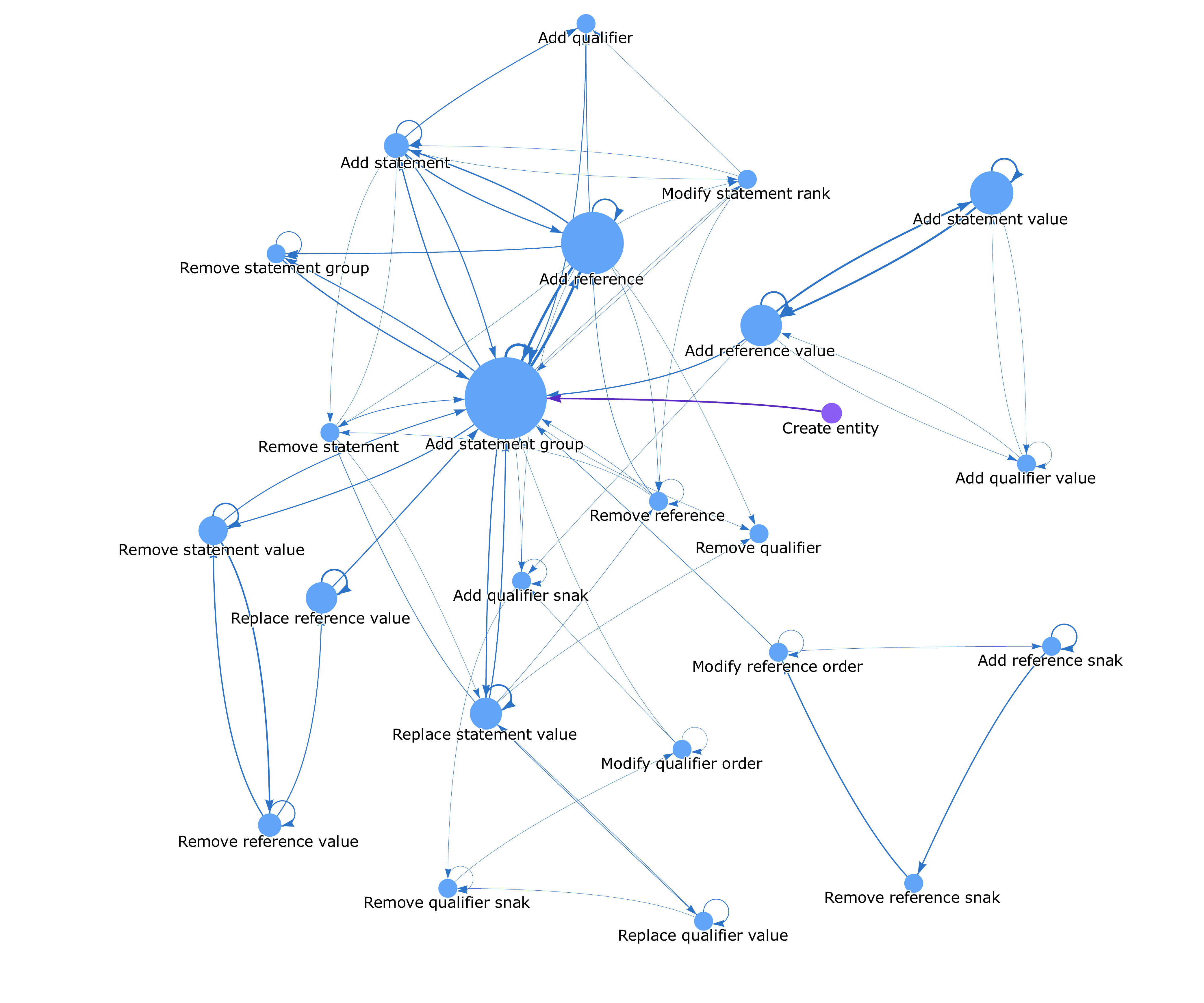}
\caption{Most common transitions from one edit operation to another in the entities lifetime. The purple circle represents the start in the life cycle of an entity, and blue circles represent different operations applied to the entity during its life cycle.}
\label{fig:states-graph}
\end{figure*}

Figure \ref{fig:states-graph} shows a graph of transitions from one edit operation applied to an entity to the next one. Edge weights represent the number of transitions from one operation to another, while node size represents the total for each state --with higher weights representing a larger count. These transition and state counts were accumulated across every entity in the dataset. In order to better illustrate the most common transitions and reduce noise from the graph, those transitions with a count lower than 10\% of the total count of outgoing transitions from a state were removed.

We can see how the central operation in this graph is adding a statement group to an entity, followed by adding references to a claim. These are the most prominent operations applied to entities in Wikidata. There is also a group of three operations in the top right corner of the graph, which represents adding single values to the entity: either a single statement, reference, or qualifier. Both behaviors are expected, since a common practice in Wikidata is to add appropriate references and qualifiers to a statement that has been added.

In general, replacements or removals are not as common as addition operations. However, these operations are usually connected between them (i.e., a replacement or removal operation is usually followed by another replacement or removal). While adding a statement may not imply adding a reference or qualifier to it, removing a statement implies the removal of its qualifiers and references. Therefore, some connections between a removal state and the next one are expected.

Finally, changing the rank or order of statements, references, or qualifiers is also an infrequent operation within Wikidata.

\subsection{Edit information}
\begin{figure*}[ht]
\centering
\includegraphics[width=0.85\textwidth]{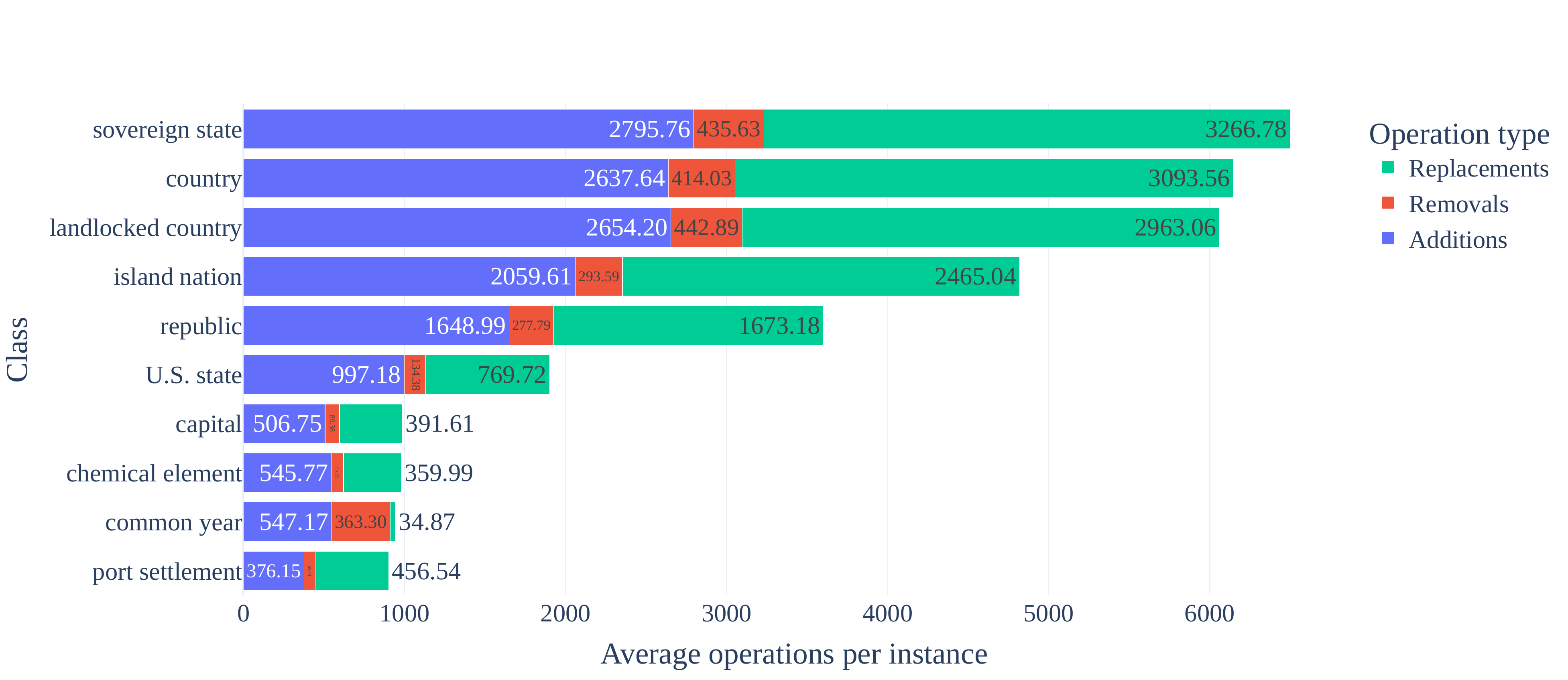}
\caption{Top 10 classes with the most number of operations performed on its instances.}
\label{fig:ops-classes}
\end{figure*}

\paragraph{\textbf{Operations applied to the most important classes}}
Figure \ref{fig:ops-classes} shows the top 10 classes with the highest average number of operations performed on its instances. These frequencies are split into additions, removals, and replacements. We can see how in the classes that have a higher operation count replacements are the most prominent operation (over additions and removals). As the number of operations keeps decreasing, replacements are less common and additions become the most common operation. The ratio between additions and removals can be higher or smaller depending on the class, which could reveal of the difficulty of reaching a consensus on the content of its instances.

Another interesting insight from this chart is that some of the most important classes based on their ClassRank score do not have a high number of average operations per instance. In particular, the 2 most important classes (human and taxon) do not appear in the top 10 classes with the most operations. One possible explanation for this could be that instances of these classes do not have as much information to be added as instances of other classes. Another possibility could be that these classes contain a high number of instances that are not edited as often, so the average number of operations is not as high as in other classes.

\begin{figure*}
    \centering
    \begin{subfigure}[b]{0.475\textwidth}
        \centering
        \includegraphics[width=\textwidth]{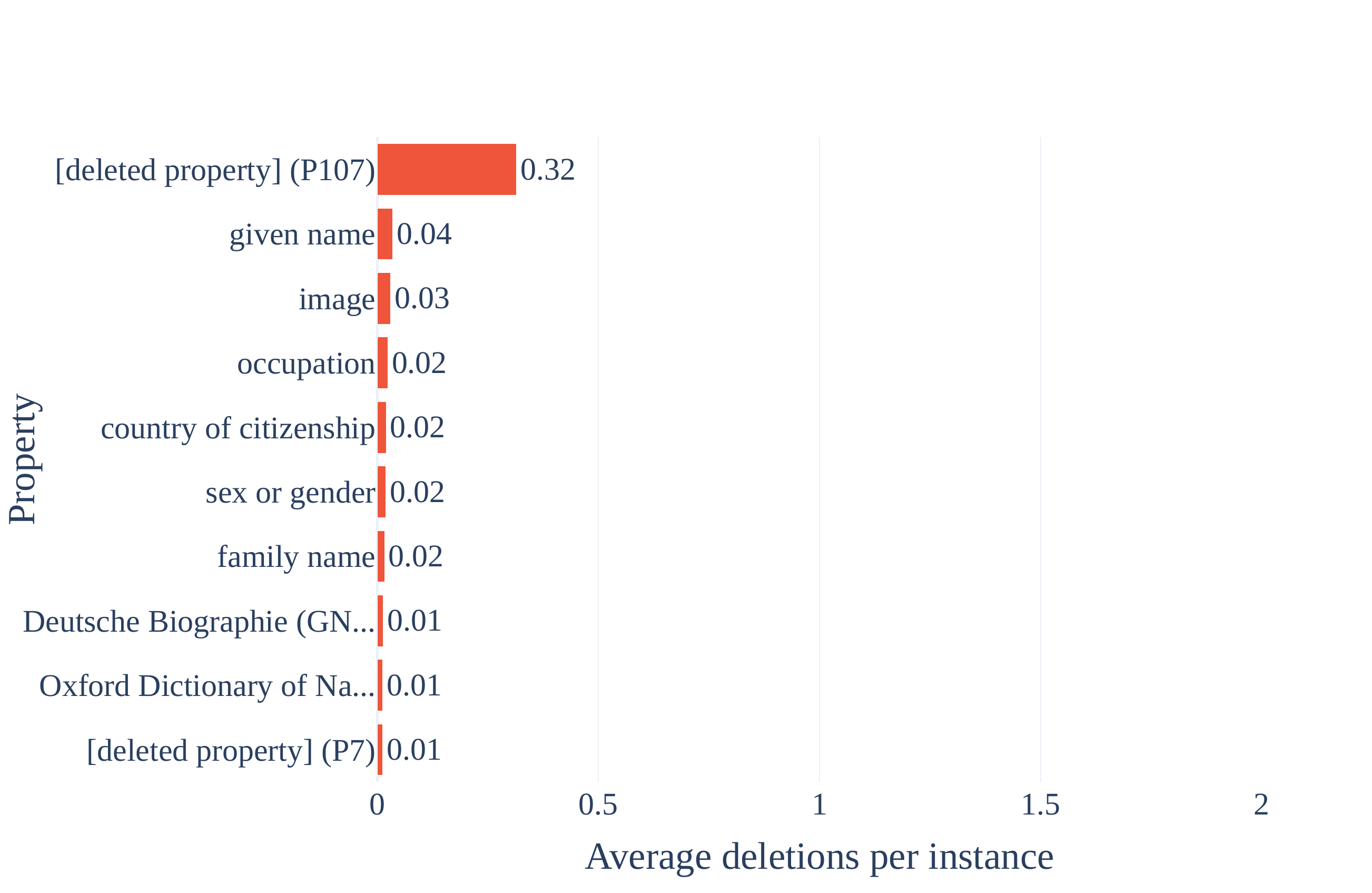}
        \caption{Class: Human}
        \label{fig:deletions-human}
    \end{subfigure}
    \hfill
    \begin{subfigure}[b]{0.475\textwidth}
        \centering
        \includegraphics[width=\textwidth]{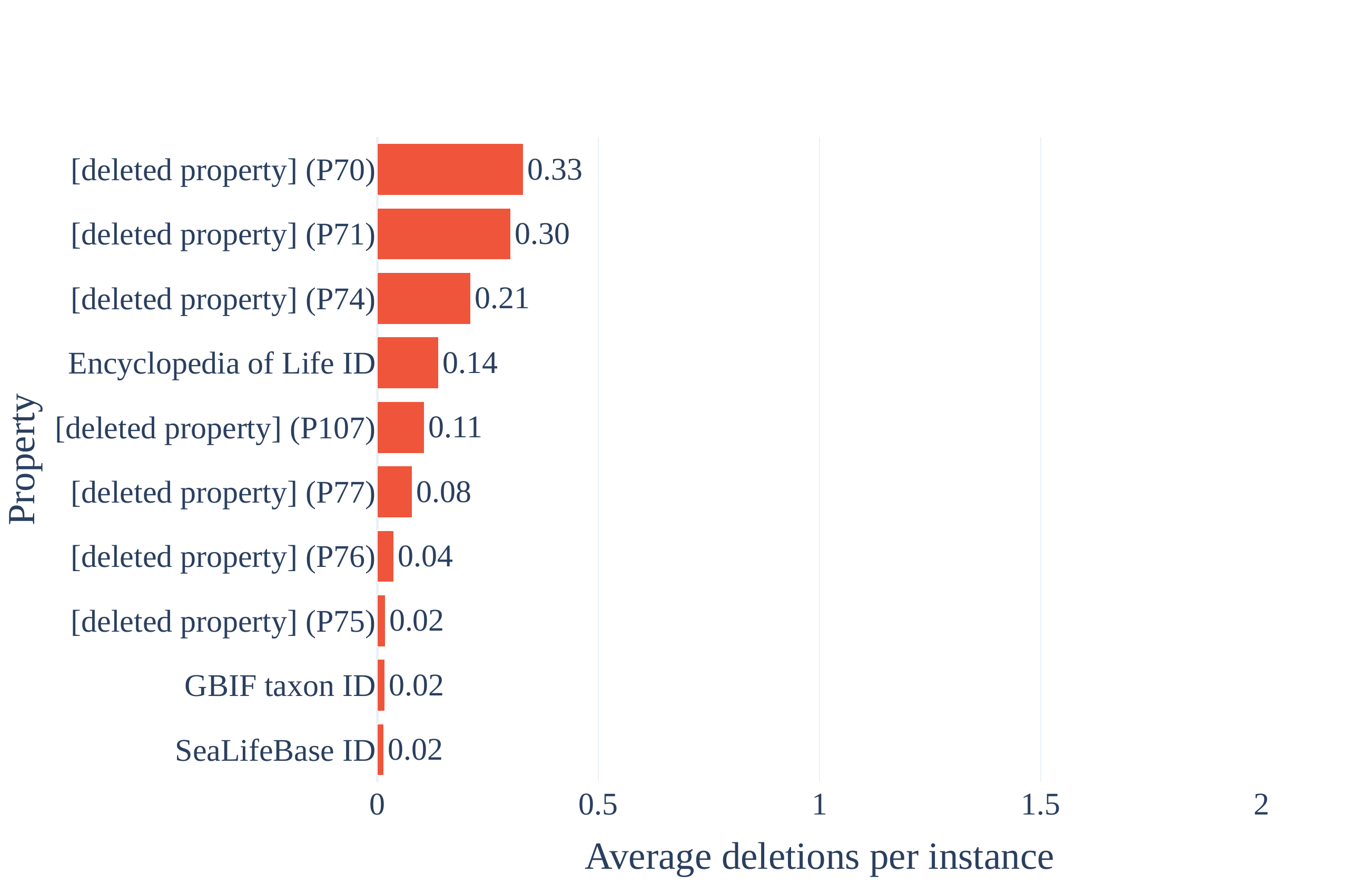}
        \caption{Class: Taxon}
        \label{fig:deletions-taxon}
    \end{subfigure}
    \hfill
    \begin{subfigure}[b]{0.475\textwidth}
        \centering
        \includegraphics[width=\textwidth]{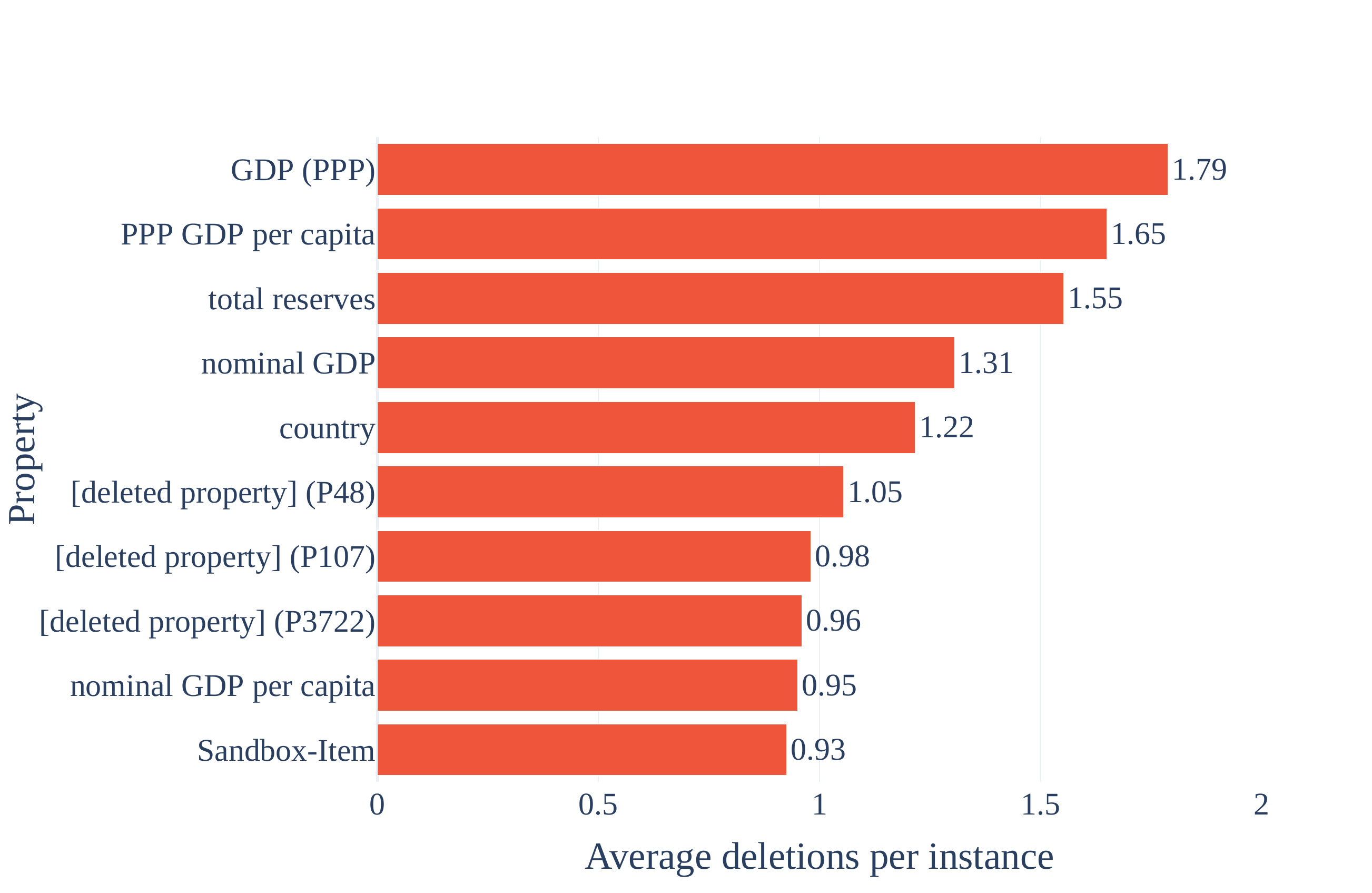}
        \caption{Class: Sovereign State}
        \label{fig:deletions-state}
    \end{subfigure}
    \hfill
    \begin{subfigure}[b]{0.475\textwidth}
        \centering
        \includegraphics[width=\textwidth]{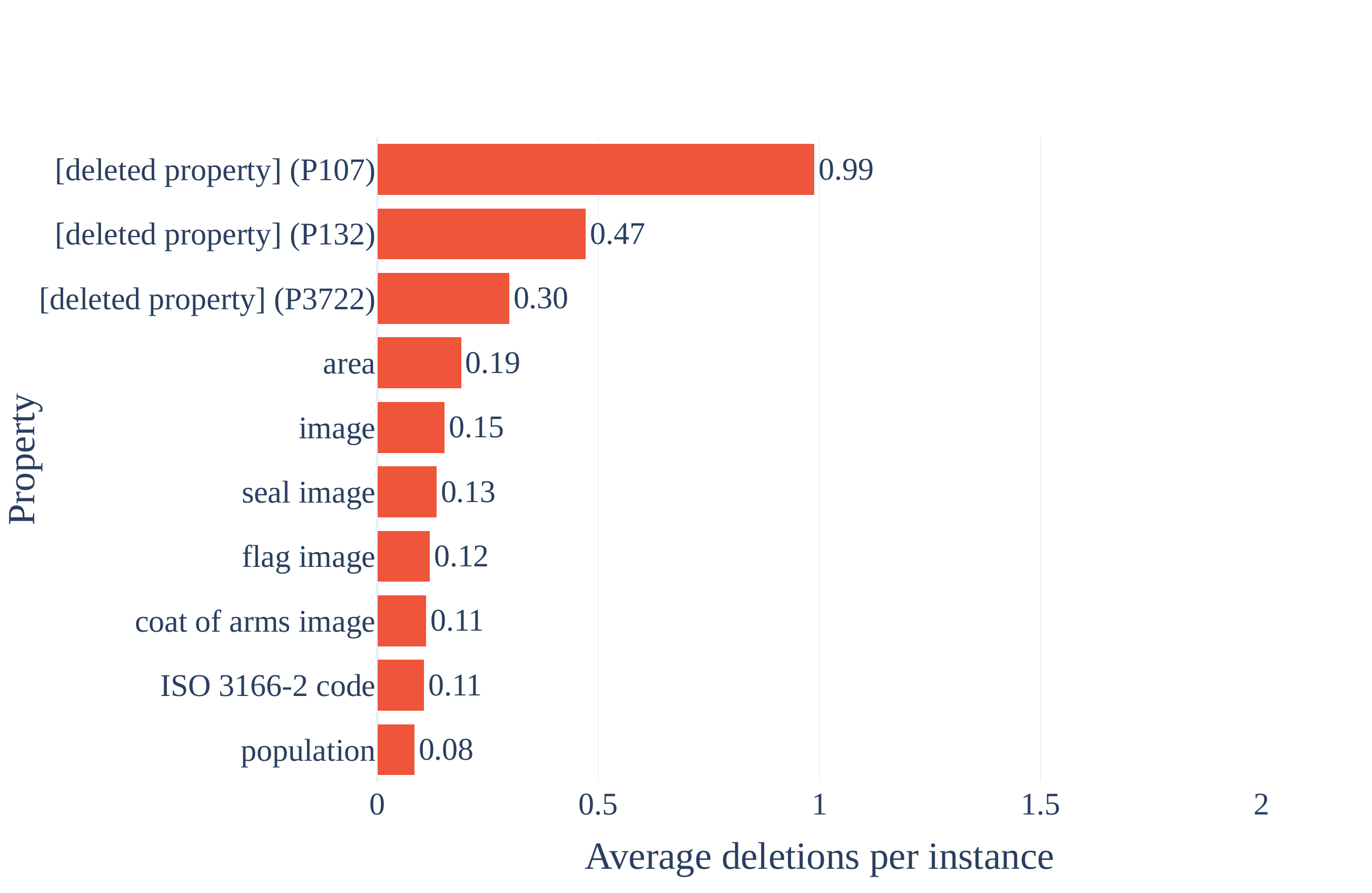}
        \caption{Class: Big City}
        \label{fig:deletions-bigcity}
    \end{subfigure}
    \caption{Most deleted properties of each class.}
    \label{fig:deletions-classes}
\end{figure*}

\paragraph{\textbf{Most frequently removed properties of a class}}
Figure \ref{fig:deletions-classes} shows the most removed properties of 4 classes from the dataset: \textit{human}, \textit{taxon}, \textit{sovereign state}, and \textit{big city}. Properties with the name `\textit{[deleted property](PXX)}' refer to properties that were removed from Wikidata, for which the original label cannot be retrieved\footnote{The process of creating and removing properties from Wikidata is more complex than creating entities or adding statements to them. It can only be done by users who have the `\textit{Property Creator}' role, and involves creating a formal proposal and reaching a consensus over the proposal.}.

We can see in the figure some differences in the removal behavior for different classes. For some classes, like `\textit{taxon}' and `\textit{big city}', the most removed properties are properties that were directly removed from Wikidata. These properties do not reflect the level of consensus in the class, since they are removed completely from Wikidata at an ontological level. The remaining properties have a lower deletion rate, indicating that the properties used in these classes are overall stable and rarely change.

Other classes, like `\textit{sovereign state}', have a high number of removals across many properties --which are still existing in Wikidata. This means that, in general, instances of that class have a set of properties that either lead to controversy or are evolving naturally along the lifetime of the instance.

Finally, the `\textit{human}' class only has a property with more than $0.3$ deletions per instance, which was removed from Wikidata. The rest of the properties have a really low deletion rate. This follows the same trend that we have seen when analyzing the number of operations per class, showing that although the human class is the most important one in Wikidata --based on ClassRank-- it does not have a high average number of operations per instance.

A deep understanding of these results across every class could be beneficial to analyze which properties are more controversial and which ones do not experience as many deletions. This information could be used by recommendation systems to aid editors by signaling those properties that are more controversial and need special consideration when used.

\subsection{Conflict in Wikidata}
Our last set of analyses of the edit history dataset revolves around the idea of divisiveness in Wikidata. We have analyzed the edits dataset to detect cases where an \textit{edit war} between users occurs when editing an entity\footnote{The interested reader should consult \cite{yasseri2012dynamics} for more information on edit wars, albeit in Wikipedia, not in Wikidata.}. In our case we have defined an \textit{edit war} as a succession of edits in which a value for a property is added, the value is removed or replaced by another different value in a later revision, and finally the original value is added again to the entity. Certainly, such a definition does not only cover the case of a conflict within the community on the value that an entity should have for a given property, but also acts of vandalism that are eventually reverted.

\begin{figure}
\centering
\includegraphics[width=0.50\textwidth]{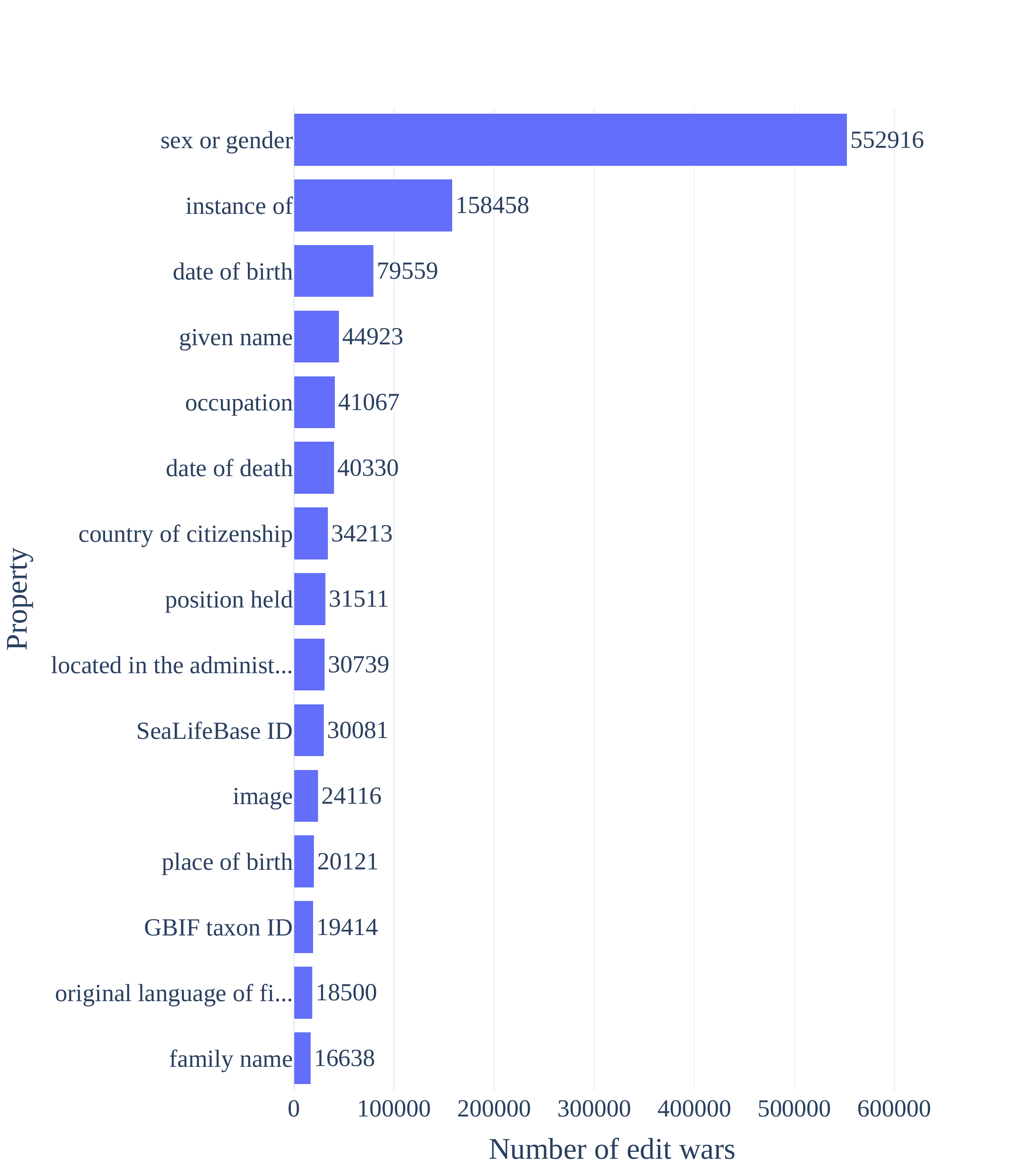}
\caption{Top 15 properties with the highest amount of edit wars.}
\label{fig:conflict-properties}
\end{figure}

Figure \ref{fig:conflict-properties} shows the properties with the most amount of edit wars in the dataset. The property with the highest edit war count was `\textit{sex or gender}', with a notable difference when compared to other properties. Typing entities with the `\textit{instance of}' property is the next case with the highest amount of edit wars. The remaining properties are more domain-specific and have a considerably lower number of edit wars when compared to the top 2.

\begin{figure}
\centering
\includegraphics[width=0.50\textwidth]{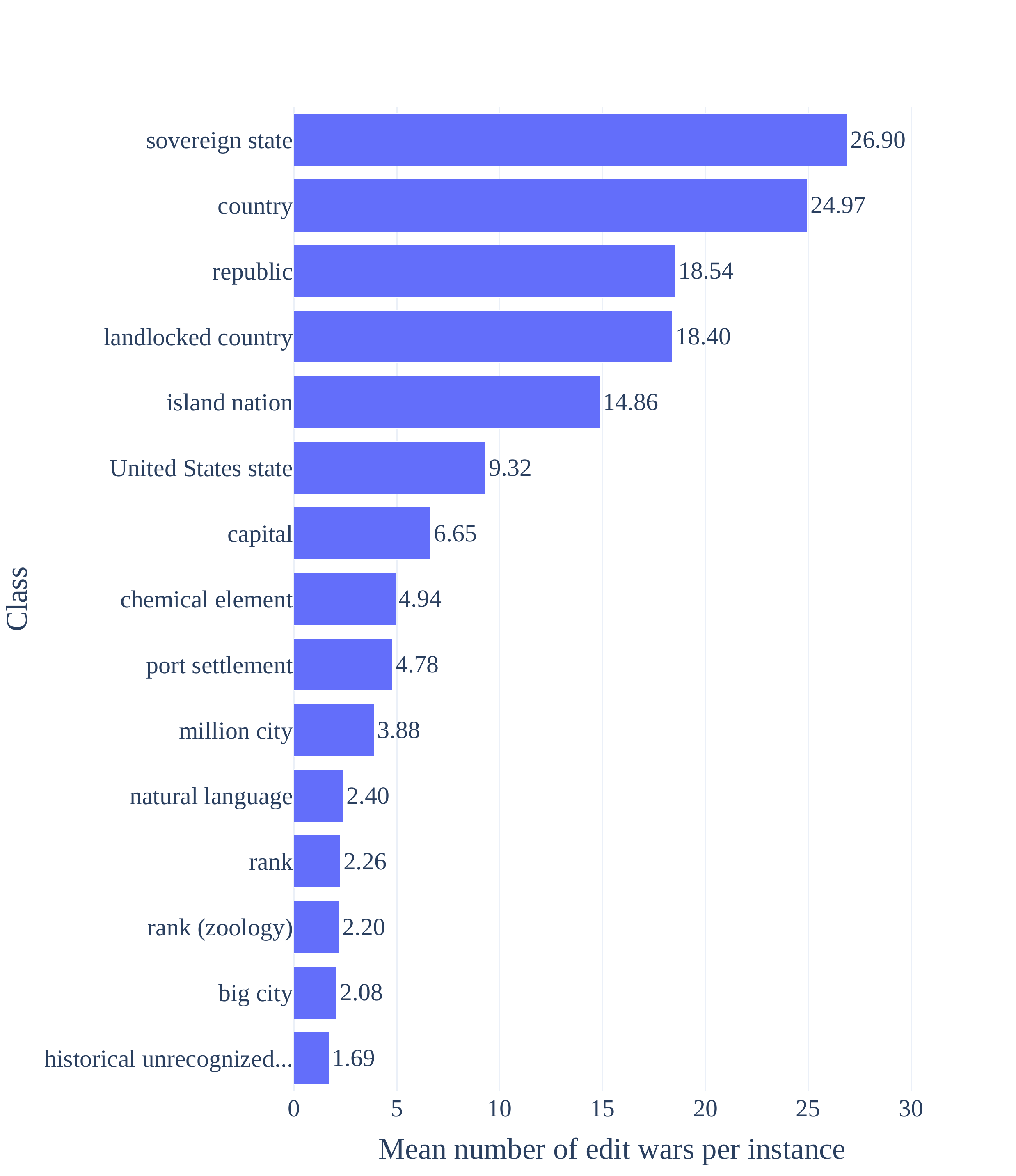}
\caption{Top 15 classes with the highest amount of edit wars.}
\label{fig:conflict-classes}
\end{figure}

In figure \ref{fig:conflict-classes} we can see the classes with the highest mean amount of edit wars across its instances. Most edit wars occur in classes related to territories, with only some exceptions not belonging to that domain (e.g., chemical element, natural language...). This follows the same pattern we have seen previously, with the `\textit{sovereign state}' class having a high number of deletions and replacements of property values.

\section{Type prediction leveraging edit data}
\label{sec:type-pred}
In this section, we explore a potential sub-field within knowledge graph refinement where this edit history data could be exploited: type prediction in knowledge graphs. More specifically, our objective is predicting possible values of the \textit{instance of} (\textit{P31}) property of a given entity. In the following sections, we will explore our approach and the results obtained.

\subsection{Proposed approaches}
\label{sec:type-pred-proposals}
In this paper we analyze the impact of leveraging edit history information in knowledge graph embedding models. We have decided to employ these models since they are currently used in type prediction and other knowledge graph refinement tasks.

Since these models usually employ a \textit{`static'} version of a graph (i.e., without including any information about the editions that the graph underwent until reaching its final version), our objective is to include information from the edit history of the graph to obtain an improvement in their performance. To do so, we propose the following two techniques in our experiments: using the removed triples from the graph to enrich negative sampling methods, and labeling each triple that was added or removed from the graph to train a supervised model.

\subsubsection{Negative sampling enriched by edit history information}
Knowledge graph embedding models perform a process called \textit{negative sampling} when generating the embeddings, where negative triples --corruptions-- are generated. Those corruptions are used during the training process of the embeddings as examples of negative triples. This generation of corruptions is usually carried out randomly, under the premise that the total space of possible triples ($|\mathcal{S} \times \mathcal{P} \times \mathcal{O}|$) is much greater than the one of true triples, and therefore the possibility of introducing false negatives is small.

However, in the revision history of an entity we have potentially useful information to determine which corruptions could be generated: triples that the community had considered invalid sometime along the life cycle of a given entity. We can draw an analogy of this idea to the task of training an automatic spell checker. One option could be to artificially generate typos on the set of training words by randomly replacing, removing, or adding one character to the word (basic negative sampling approach). However, a more optimal approach could be to also use as a training sample real typos made by humans on these words when writing on their devices (edit history based negative sampling approach).

Therefore, we propose the following set of negative sampling methods based on the edit history of a knowledge graph:

\begin{itemize}
    \item \textbf{Edit history negative sampling}. We start by selecting the set of negative triples of each triple that needs to be corrupted. This set of negative triples is obtained from all the removal operations in the knowledge graph that share the same subject and predicate of the given positive triple. For example, if we want to generate corruptions for the triple `\texttt{DouglasAdams occupation writer}', we would look into the edit history of the graph for triples that were removed having `\texttt{DouglasAdams}' as subject and `\texttt{occupation}' as predicate\footnote{In these examples the corruption of objects of each triple is shown. The corruption of subjects is analogous, where we instead select as negative triples those that share the same predicate and object of the positive triple.}. Afterward, the set of negative triples is shuffled, and the first $n$ elements are selected as the generated corruptions. If the set of negative triples contains less than $n$ elements, the remaining corruptions are generated randomly, following the traditional process. A variation of this technique consists of generating the set of negative triples without including those triples that were the subject of edit wars. In the following experiments, we will refer to models using this type of negative sampling techniques with the terms \textbf{`edit hist'} y \textbf{`edit hist (no edit wars)'}, with the former including edit wars in the set of negative triples while the later omits them.
    \item \textbf{Inverse edit history negative sampling}. This method is based on the opposite hypothesis with respect to the previous one: triples that were removed by the community represent a conflict of opinion about the value of a property, and they do not entail that its value is necessarily invalid. Therefore, these removed triples should not be added to the corruptions of the model since they are triples that could be considered valid or, at least, partially related to the entity that is being corrupted. For this reason, although we start from the same set of negative triples explained previously, in this case the objective is to avoid the generation of any corruption that belongs to that set. Corruptions are generated randomly, following the traditional approach, but if the generated corruption belongs to the set of negative triples the corruption is generated again randomly until it does not belong to it.
\end{itemize}

The pseudo-code of the corruption fetching process and both negative sampling approaches is shown in \autoref{alg:corruption-fetching} and \autoref{alg:negative-sampling} respectively.

\begin{algorithm*}
	\DontPrintSemicolon
	\KwIn{}
	\myinput{$t = (s, p, o)$ Triple that needs to be corrupted.}
	\myinput{$Op^- = \{t_1, t_2 ... t_m\}$ Set of removal operations of the knowledge graph.}
	\myinput{$omit\_edit\_wars \in \{True, False\}$  Whether edit wars should be omitted or not.}
	\vspace{0.2cm}
	\Variables{}
	\myinput{$F \subset Op^-$ Set of all filtered corruptions.}
	\vspace{0.2cm}
	\KwOut{}
	\myinput{$T_{neg} \subset F$ Set of negative samples of $t$.}
	\vspace{0.2cm}
	\Begin{
		$F \longleftarrow \emptyset$\;
		\For{$t'=(s', p', o') \in Op^-$}{
			\If{$omit\_edit\_wars = False$ or not $HasEditWars(t')$}{
				$F \longleftarrow F \cup \{t'\}$ \;
			}
		}
		$T_{neg} \longleftarrow \{(s', p', o') \in F : s' = s, p' = p\}$\;
		\Return $T_{neg}$
	}
	\caption{Fetching of corruptions from the edit history pseudo-code}
    \label{alg:corruption-fetching}
\end{algorithm*}

\begin{algorithm*}
    \DontPrintSemicolon
    \KwIn{}
    \myinput{$n \in \mathbb{N}$: Number of negative samples to be generated per triple.}
    \myinput{$Op^-$ Set of removal operations of the knowledge graph.}
	\myinput{$T=\{t_1, t_2, ..., t_m\}$ Set of triples to be corrupted.}
    \myinput{$omit\_edit\_wars \in \{True, False\}$  Whether edit wars should be omitted or not.}
    \vspace{0.2cm}
	\Variables{}
	\myinput{$T_{neg}$ Set of negative samples of triple $t$.}
	\vspace{0.2cm}
	\KwOut{}
	\myinput{$C_{out} \subset (\mathcal{I} \cup \mathcal{B}) \times \mathcal{I} \times (\mathcal{I} \cup \mathcal{L} \cup \mathcal{B})$ Generated corruptions.}
	\vspace{0.2cm}
    \SetKwFunction{FMain}{Main}
    \SetKwProg{Fn}{Function}{:}{}
    \Fn{InverseNegativeSampling({$T$, $Op^-$, $omit\_edit\_wars$, $n$})}{
      $C_{out} \longleftarrow \emptyset$\;
		\For{$t=(s, p, o) \in T$}{
		    $T_{neg} \longleftarrow FetchCorruptions(Op^-, t, omit\_edit\_wars)$\;
		    \For{$i=1:n$}{
    			\Do{$t' \notin T_{neg}$} {
    				$t' \longleftarrow RandomCorruption(t)$\;
    			}
			
			    $C_{out} \longleftarrow C_{out} \cup {t'}$
			}
		}
		\Return $C_{out}$\;
    }
    \;
    \SetKwProg{Pn}{Function}{:}{\KwRet}
    \Pn{EditHistoryNegativeSampling({$T$, $Op^-$, $omit\_edit\_wars$, $n$})}{
		$C_{out} \longleftarrow \emptyset$\;
		\For{$t=(s, p, o) \in T$}{
		    $T_{neg} \longleftarrow FetchCorruptions(Op^-, t, omit\_edit\_wars)$\;
			\While{$|T_{neg}| < n$} {
				$T_{neg} \longleftarrow T_{neg} \cup RandomCorruption(t)$
			}
			
			$T_{neg} \longleftarrow Shuffle(T_{neg})$\;
			$C_{out} \longleftarrow C_{out} \cup \{t_1, t_2 ... t_n : t_i \in T_{neg}\}$
		}
		\Return $C_{out}$
    }
    
    \caption{Negative sampling methods pseudo-code}
    \label{alg:negative-sampling}
\end{algorithm*}

\subsubsection{Transformation of non-supervised task to a supervised one}
While the generation of knowledge graph embeddings is a non-supervised technique where it is not necessary to label any triple from the knowledge graph, having information about the edit history of the graph allows us to convert this task into a supervised one.

To be more precise, from the set of addition operations $Op^+$ and the set of removal operations $Op^-$, we can assign a negative label $y=0$ to each triple $t \in Op^-$, while assigning a positive label $y=1$ to each triple $t \in Op^+$. With this approach, we can train a predictive model from the embeddings representation of each triple and their corresponding label obtained from the edit history of the knowledge graph. In our experiments the input of the classifiers will be the concatenation of the subject, predicate, and object embeddings. In order to obtain a ranking of triples for evaluation purposes, we have ordered the results by the log probability given by the classifier of the results being positive.

\subsection{Dataset}
\label{sec:type-pred-dataset}
Knowledge graph embedding models work on a graph version of a dataset, while our dataset indexed on MongoDB that has been previously explored is stored in JSON format. Therefore, to run these models it was necessary to transform that data into RDF.

The first step of that process is to transform the `static' JSON content of each entity to RDF. To do so, we transformed just the \textit{simple statements} of every entity, ignoring complex values like qualifiers and references. Since the JSON dataset indexed in MongoDB is composed of millions of triples, a subset of this data was selected to be converted to RDF. This subset has 300 randomly selected instances from each class indexed in MongoDB, resulting in 390.000 triples in the final graph.

We also created an additional RDF dataset containing the `dynamic' information of each entity (i.e., the edit history information). This dataset represents each revision made to an entity, plus the atomic operations that compose the revision. It is possible to build the static RDF graph by applying each operation on the dynamic version, following the same principles as with the JSON dataset.

Finally, both datasets (dynamic and static) were split into training, validation, and test sets. Every entity is maintained on each set when making the divisions, but in the training dataset we include the state of every entity until a time step $t_i$, in the validation set their new state from a time step $t_i$ until $t_{i+n}$, and in the test set its new state since time step $t_{i+n}$ until now. In our particular case, the training set includes the first 70\% revisions of each entity (in chronological order), the validation set includes an additional 15\% (between the 70\% and 85\% revisions), and the test set contains the last 15\% of revisions.

\subsection{Training}
\paragraph{\textbf{Unsupervised models}}
We have selected 3 knowledge graph embedding models to evaluate the impact of our proposed negative sampling techniques: RotatE \cite{sun2019rotate}, TransE \cite{bordes2013translating}, and MuRE \cite{balazevic2019multi}. Those models were selected based on the large-scale benchmarks done by Ali et al. under the PyKEEN framework \cite{ali2021bringing}. We have chosen those with a good performance under the Yago dataset\footnote{Due to it being the dataset more closely related to the one used in this paper.} and that can scale to large dataset sizes without requiring large computational resources.

Hyperparameters were optimized for each model using the training and validation sets. Since the total space of parameter combinations was too big to be completely covered, a random search was made to find the best parameters. Each model parameters were optimized for 80h, for a total of 240h (10 days). Table \ref{tbl:hyperparams-unsupervised} shows the final set of optimized parameters found for each model. Regarding parameters specific to each model, TransE had an optimized parameter $scoring\_fct\_norm=2$ and MuRE's best value for the $l_p$ norm parameter was $2$.

\begin{table*}
	\caption{Optimized hyperparameters of each unsupervised model}
	\label{tbl:hyperparams-unsupervised}
	\centering
	\begin{tabular}{l r r r r r}
	\toprule
    	Model & Embedding dim. & Num. epochs & Batch size & Learning rate & Num. negatives \\
	\midrule
        \textbf{RotatE} & 768 & 13 & 64 & 0.009 & 5 \\
        \textbf{TransE} & 64 & 10 & 521 & 0.024 & 7 \\
        \textbf{MuRE} & 150 & 21 & 256 & 0.088 & 28 \\
	\bottomrule
	\end{tabular}
\end{table*}

To make the most out of the available data in the dataset, each combination of model and negative sampler was then trained on the union of both the training and validation sets. That is, the final training set of each model contains the 85\% first revisions of each entity, while the last 15\% of revisions are reserved for the test set.

\paragraph{\textbf{Supervised model}}
For the supervised version of this task we have first trained RDF2Vec embeddings \cite{ristoski2016rdf2vec} on the dataset, and fed the embeddings to a Random Forest classifier with the labeled data.

The same optimization and training process done with the unsupervised models was followed with the supervised approach. In this case, we have also optimized the parameters of both the RDF2Vec and the Random Forest models, performing a random search over a space of parameters for each model. The final set of parameters found for the RDF2Vec model was a vector size of $50$, with $50$ training epochs, a maximum depth of $5$, and $50$ walks generated. Regarding the random forest classifier, the set of parameters found was $75$ for the number of trees, a minimum of $2$ samples to perform a split, and the $gini$ criterion to perform splits.

\subsection{Evaluation}
To measure the performance of each model in the test set we have used the following metrics: \textbf{mean rank (MR)}, \textbf{mean reciprocal rank (MRR)}, and \textbf{hits@1}, \textbf{5}, and \textbf{10}. We selected those entities which had a new class added to the test set (i.e., a new value for the \textit{instance of} property), and run each model to get the ranked list of class suggestions for the entity. We have followed the `filtered setting' proposed by Bordes et al. in the evaluation process by removing those triples suggested by the model which had already appeared in the train set (\textit{true triples}) \cite{bordes2013translating}. The final test sample is made of 4193 triples, which are the triples from the test dataset containing the \textit{instance of (P31)} property.

\subsection{Results}
\subsubsection{Negative sampling techniques}
\begin{figure*}
\centering
\includegraphics[width=0.975\textwidth]{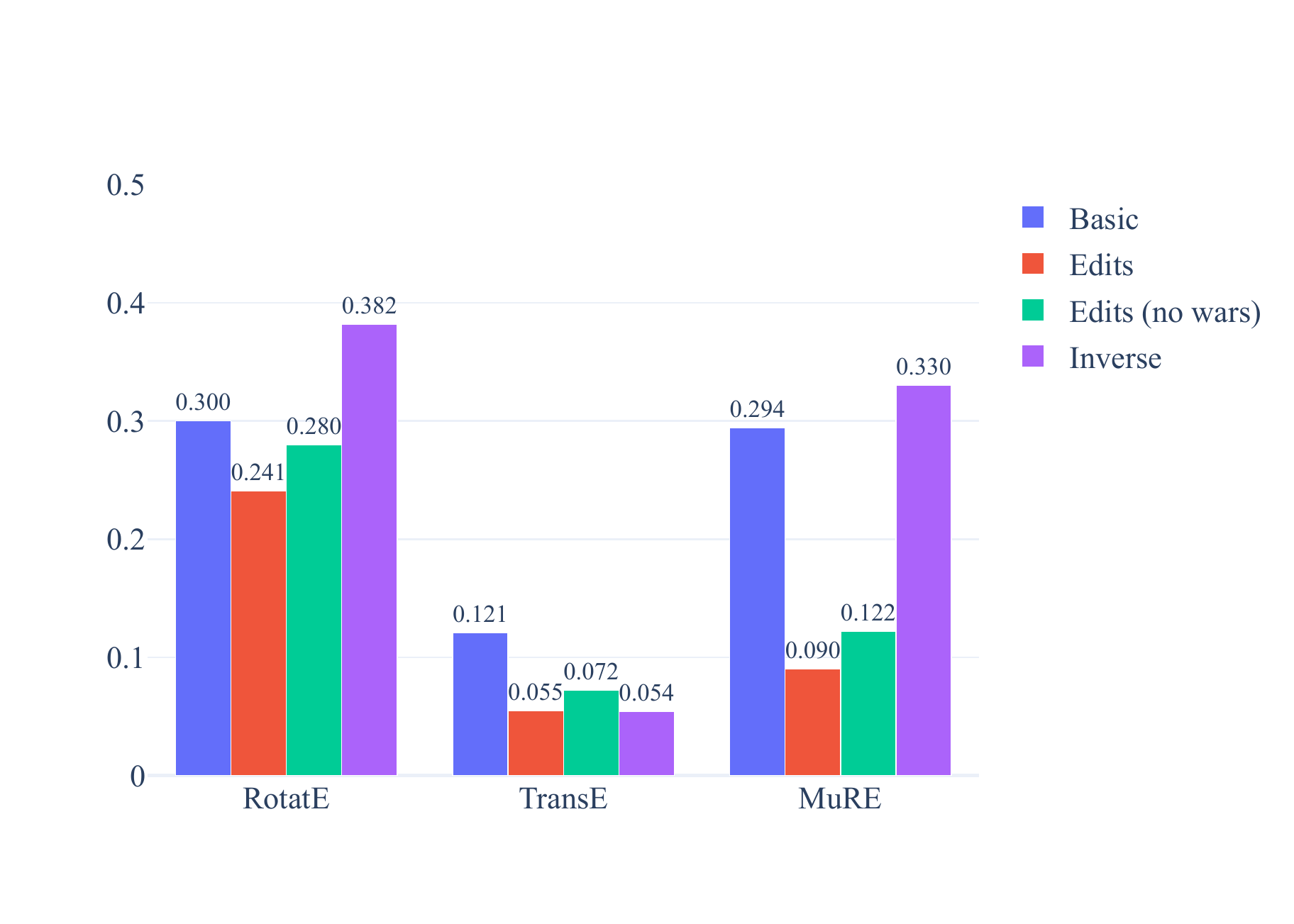}
\caption{Hits@5 results of each unsupervised model.}
\label{fig:hits-at-5}
\end{figure*}

The results obtained for each combination of unsupervised model and negative sampling technique are shown in table \ref{tbl:evaluation-unsupervised}. Results for the \textit{hits@5} measure across every model combination are graphically shown in figure \ref{fig:hits-at-5}.

{
\begin{table*}

\renewcommand{\arraystretch}{1.35}
	\caption{Evaluation results of unsupervised models. `\textit{basic}' represents the basic negative sampler currently used in knowledge graph embedding models; `\textit{edits}' and `\textit{edits (no w.)}' refer to the proposed negative sampling techniques that generate corrupt triples from removals of the edit history, considering or omitting edit wars respectively; and `\textit{inv.}' refers to the proposed inverse edit history negative sampling, where triples that were removed from the knowledge graph cannot be generated as corruptions.}
	\label{tbl:evaluation-unsupervised}
	\centering
	\begin{tabular}{l | r r >{\centering\arraybackslash}p{1.15cm} r | r r >{\centering\arraybackslash}p{1.15cm} r | r r >{\centering\arraybackslash}p{1.15cm} r}
	\toprule
	    & \multicolumn{4}{c|}{RotatE} & \multicolumn{4}{c|}{TransE} & \multicolumn{4}{c}{MuRE} \\
	    & basic & edits & edits (no w.) & inv. & basic & edits & edits (no w.) & inv. & basic & edits & edits (no w.) & inv.\\
	\hline
        \textbf{MR} & 10819 & 21268 & 23577 & \textbf{7197} & 3204 & \textbf{2385} & 2756 & 2595 & 527 & 4650 & 3385 & \textbf{447} \\
        \textbf{MRR} & 0.177 & 0.175 & 0.211 & \textbf{0.260} & \textbf{0.091} & 0.043 & 0.055 & 0.046 & 0.204 & 0.063 & 0.085 & \textbf{0.237}\\
        \textbf{hits@1} & 0.086 & 0.109 & 0.146 & \textbf{0.163} & \textbf{0.050} & 0.015 & 0.028 & 0.020 & 0.115 & 0.018 & 0.031 & \textbf{0.144}\\
        \textbf{hits@5} & 0.300 & 0.241 & 0.280 & \textbf{0.382} & \textbf{0.121} & 0.055 & 0.072 & 0.054 & 0.294 & 0.090 & 0.122 & \textbf{0.330}\\
        \textbf{hits@10} & 0.382 & 0.308 & 0.339 & \textbf{0.445} & \textbf{0.164} & 0.091 & 0.097 & 0.083 & 0.385 & 0.148 & 0.210 & \textbf{0.422}\\
	\bottomrule
	\end{tabular}
\end{table*}
}

We performed McNemar tests over each combination of models to prove whether there are statistically significant differences ($pvalue \leq 0.01$) between the results obtained \cite{dietterich1998approximate}. To build the contingency table we measured the accuracy (hits@1) of each model being compared, storing those cases when both models guessed correctly the new class of an entity, those cases when both models failed to guess the class, and the cases when one of the models guessed the class correctly while the other did not. Therefore, each comparison will measure whether both models have a similar proportion of errors in the test set. Since we are interested in the differences between each negative sampler, we only performed comparisons between the different negative samplers of each model type.

Results of running the McNemar test over each pair of negative samplers show the following statistically significant differences for the RotatE model: \textit{basic} against \textit{edits} $[\chi^2(1)=15.00, \; pvalue=1 \cdot 10^{-4}]$; \textit{basic} against \textit{edits (no wars)} $[\chi^2(1)=79.27, \; pvalue=5.4 \cdot 10^{-19}]$; \textit{basic} against \textit{inverse} $[\chi^2(1)=144.27, \; pvalue=3.1 \cdot 10^{-33}]$; \textit{edits} against \textit{edits (no wars)} $[\chi^2(1)=39.75, \; pvalue=2.8 \cdot 10^{-10}]$;  \textit{edits} against \textit{inverse} $[\chi^2(1)=72.20, \; pvalue=1.9 \cdot 10^{-17}]$.

With regards to the TransE model, we obtain the following statistically significant differences: \textit{basic} against \textit{edits} $[\chi^2(1)=72.90, \; pvalue=1.3 \cdot 10^{-17}]$; \textit{basic} against \textit{edits (no wars)} $[\chi^2(1)=25.44, \; pvalue=4.5 \cdot 10^{-7}]$; \textit{basic} against \textit{inverse} $[\chi^2(1)=54.37, \; pvalue=1.6 \cdot 10^{-13}]$; \textit{edits} against \textit{edits (no wars)} $[\chi^2(1)=16.66, \; pvalue=4.4 \cdot 10^{-5}]$;.

Regarding the MuRE model, there is a statistically significant difference between every negative sampler, with the following detailed results: \textit{basic} against \textit{edits} $[\chi^2(1)=274.69, \; pvalue=1 \cdot 10^{-61}]$; \textit{basic} against \textit{edits (no wars)} $[\chi^2(1)=206.86, \; pvalue=6.6 \cdot 10^{-47}]$; \textit{basic} against \textit{inverse} $[\chi^2(1)=20.48, \; pvalue=6.0 \cdot 10^{-6}]$; \textit{edits} against \textit{edits (no wars)} $[\chi^2(1)=12.69, \; pvalue=3.6 \cdot 10^{-4}]$;  \textit{edits} against \textit{inverse} $[\chi^2(1)=393.26, \; pvalue=1.6 \cdot 10^{-87}]$;  \textit{edits (no wars)} against \textit{inverse} $[\chi^2(1)=307.65, \; pvalue=7.1 \cdot 10^{-69}]$.

The remaining combinations of negative samplers do not show statistically significant differences ($pvalue > 0.01$), with the following detailed results: RotatE \textit{edits (no wars)} and RotatE \textit{inverse} $[\chi^2(1)=6.01, \; pvalue=0.014]$; TransE \textit{edits} and TransE \textit{inverse} $[\chi^2(1)=2.58, \; pvalue=0.10]$; TransE \textit{edits (no wars)} and TransE \textit{inverse} $[\chi^2(1)=5.80,$ $\; pvalue=0.016]$.

In order to report the impact of our proposed negative samplers we also follow the evaluation reporting approach proposed by Sparck-Jones \cite{jones1974automatic}: performance differences are regarded as \textit{non-relevant} if they are lower than $5\%$; as \textit{noticeable} if they are between $5-10\%$; and as \textit{material} if they are greater than $10\%$.

We can observe from the results that the use of the inverse edit history negative sampling technique has a positive effect on both the RotatE and MuRE models across every measure, having a \textit{material} and \textit{noticeable} impact respectively. However, with the TransE model this is not the case, with the performance of the basic negative sampler improving that of the proposed negative samplers across almost every measure. The only exception is the mean rank measure, where the basic negative sampler has on average a $24.68\%$ higher mean rank score than the proposed approaches. It is important to note that the TransE model results are in general considerably worse than the RotatE and MuRE ones, regardless of the negative sampler being used. This could be due to not finding an optimal set of parameters for the model, or to the model having a worse performance in this dataset with respect to the other alternatives. In general, using the proposed inverse edit history negative sampler seems to influence positively in the results, improving the performance of 2 out of the 3 models being evaluated and obtaining the 2 best results in terms of MRR and hits@k.

On the other hand, an improvement over the basic approach is not as clear in the case of the edit history negative samplers. If we consider the RotatE model, both edit history negative samplers still perform better than the basic approach in terms of hits@1, but the improvements in performance are not as high as in the case of the inverse negative sampler. However, in the remaining models the performance of the edit history negative samplers is worse than the basic sampling approach across almost every metric. This could be attributed to the fact that, in some cases, a value that was removed from an entity may actually be true --or at least, partially related to the domain of the entity being affected.

We are going to illustrate this with a real example from the dataset: the entity representing \textit{MAN Truck \& Bus}\footnote{\url{https://www.wikidata.org/wiki/Q708667}} was initially an instance of `\textit{company}'\footnote{\url{https://www.wikidata.org/wiki/Q783794}} in Wikidata, but this was later removed and replaced by the `\textit{business}'\footnote{\url{https://www.wikidata.org/wiki/Q4830453}} class. Although there are some legal details that differentiate both terms, they are still really close from a conceptual point of view, especially for Wikidata contributors without expertise in business law. In cases like the one mentioned above, generating the removed triple (\texttt{MAN} \texttt{instanceof} \texttt{company}) as a corruption could add noise to the model. This could also be a reason for the good performance of the inverse negative sampling approach, since it avoids those cases.

It is also crucial to understand that there is a random component to the edit history negative sampling methods. Since we are fetching $n$ random negative triples for each triple that needs to be corrupted, in some cases we may be fetching true negative triples (e.g., an edit that removed an act of vandalism) while in other cases we may be getting negative triples with a close relation to the entity conceptually (like the MAN Truck \& Bus example mentioned above) or even a removal of a completely valid triple. This can explain some of the variability in the performance of the edit history negative samplers and leads to the need of further analysis which is addressed in the future work section.

Another interesting insight from these results is that omitting edit wars is beneficial for the performance of the edit history negative samplers across every model. This is intuitively expected, since values that cause conflict between the community should not be added --in general-- as false triples to the model. These values are more commonly associated with information that is not clear whether an entity should have or not, and therefore are not usually related to incorrect information. Furthermore, although these values were removed in the training set, since they were part of edit wars it is more probable that they will be part of the added triples in the test set. Establishing these triples as false in our negative sampling approach will decrease their suggested ranking in the testing phase, lowering the performance of the model.

\subsubsection{Supervised task}
Table \ref{tbl:evaluation-supervised} shows the results obtained by the supervised approach against each unsupervised model with its best negative sampler.

The same process was followed to determine whether there are statistically significant differences between the error rates of each model. Results of the McNemar test showed a statistically significant difference across every combination of models, with the following detailed results: RotatE against TransE $[\chi^2(1)=15.00, \; pvalue=1 \cdot 10^{-4}]$; RotatE and MuRE $[\chi^2(1)=79.27, \; pvalue=5.4 \cdot 10^{-19}]$; RotatE and RDF2Vec + Random Forest $[\chi^2(1)=15.00, \; pvalue=1 \cdot 10^{-4}]$; TransE and MuRE $[\chi^2(1)=79.27, \; pvalue=5.4 \cdot 10^{-19}]$;  TransE and RDF2Vec + Random Forest $[\chi^2(1)=79.27, \; pvalue=5.4 \cdot 10^{-19}]$; MuRE and RDF2Vec + Random Forest $[\chi^2(1)=79.27, \; pvalue=5.4 \cdot 10^{-19}]$.

As we can see in the results, the supervised approach (RDF2Vec + Random Forest) is the worst approach in terms of mean rank (\textbf{MR}) by a considerable amount, having a 34 times larger mean rank than the best approach (MuRE). This larger decrease in mean rank score with respect to other metrics is expected, since the supervised approach converts the type prediction task to a binary classification problem. That is, this model tries to maximize the probability of guessing correctly whether an entity is true or false: the ranking of true over false triples does not matter to the model.

With respect to the other metrics, its performance is better than TransE but worse than both RotatE and MuRE. This follows the same trends that were outlined when analyzing the performance of each negative sampler. Using removals from the edit history as false triples to the classifier adds a lot of noise in some cases, decreasing the overall performance of the model.

{
\begin{table*}

\renewcommand{\arraystretch}{1.35}
	\caption{Evaluation results of each best unsupervised model and the supervised model.}
	\label{tbl:evaluation-supervised}
	\centering
	\begin{tabular}{l r r r r}
	\toprule
	    & RotatE (inv.) & TransE (basic) & MuRE (inv.) & RDF2Vec + Random Forest \\
	\hline
        \textbf{MR} & 7191 & 2385 & \textbf{447} & 15752 \\
        \textbf{MRR} & \textbf{0.260} & 0.091 & 0.237 & 0.124 \\
        \textbf{hits@1} & \textbf{0.163} & 0.050 & 0.144 & 0.062 \\
        \textbf{hits@5} & \textbf{0.382} & 0.121 & 0.330 & 0.170 \\
        \textbf{hits@10} & \textbf{0.445} & 0.164 & 0.422 & 0.249 \\
	\bottomrule
	\end{tabular}
\end{table*}
}

\subsection{Reproducibility}
Experiments were executed using Python 3.8. Knowledge graph embedding models were created and trained using the PyKEEN 1.8.0 \cite{ali2021pykeen} and PyTorch 1.11.0 \cite{NEURIPS2019_9015} libraries. The supervised model was built with pyRDF2Vec 0.2.3 \cite{pyrdf2vec}, and Scikit-learn 1.0.2 \cite{scikit-learn}. A random seed of 42 was established to create the RDF dataset and train each model.

The complete code where all the experiments of this work were conducted is also available at \url{https://github.com/alejgh/wikidata-edithist-refinement}. The RDF datasets with static and dynamic serializations used in the type prediction task are also publicly available \cite{gonzalez_hevia_alejandro_2022_6613875}. We think that new research opportunities can arise by exploiting the dynamic information of an RDF dataset beyond the negative sampling techniques proposed in this paper, and both datasets could be used to measure the performance of models leveraging the edit history of a knowledge graph.

\section{Related Work}
\label{sec:rel-work}
\subsection{Collaboration dynamics of Wikidata}
Several works have tried to better understand the collaboration and edition dynamics of Wikidata. T. Steiner was one of the first authors exploring this field, by implementing a tool that provided real-time statistics about edits made in both Wikipedia and Wikidata \cite{steiner2014bots}. That author performed a quantitative analysis of the editions made to Wikidata across 3 days in 2013, focusing on the impact that bots had on the knowledge graph. Their analysis shows that --during that time frame-- most edits in Wikidata were made by bots (around $88\%$).

While the quantitative analysis of Steiner provided a better understanding of the editor distribution in Wikidata, a qualitative analysis was still needed to understand which aspects of Wikidata were edited by each contributor, and therefore get a better view of their behavior within the Wikidata ecosystem. This is one of the main contributions of the study conducted by Müller-Birn et al. in 2015 \cite{muller2015peer}. Those authors performed clustering on edit statistics computed across 1 month time frames, and those clusters were used to detect different editing roles in Wikidata (e.g., property editor, domain expert, instance creator...). Those editing roles correspond to the different collaboration patterns that exist in Wikidata.

Another point of interest within the field is understanding the path that a new editor follows to become an established contributor within Wikidata. Piscopo et al. detected that Wikidata has a low entry barrier, allowing newcomers to start making simple statement contributions right away \cite{piscopo2017wikidatians}. However, as contributors become more expert around the Wikidata ecosystem they transition from performing editions through the interface to using more advanced automatic tools. Understanding at an early stage which editors may be more engaged with editing in Wikidata was the goal of Sarasua et al. \cite{sarasua2019evolution}. They monitored the way users participated in Wikidata for 4 years, and developed predictive models to estimate the edit lifetime of a contributor reaching an F1-score of $0.9$. 

More recent works have shifted to understanding how these collaboration dynamics affect the overall quality of the knowledge graph. Piscopo et al., analyzed several edit metrics of 5,000 Wikidata items to understand these relationships \cite{piscopo2017makes}. They detected that the interaction between human and bot users is crucial to ensure the quality of Wikidata items, and that anonymous edits are instead detrimental to the quality (mainly due to vandalism). In their follow-up work, they tried to detect a correlation between a set of defined quality metrics and editing roles detected with a clustering approach \cite{piscopo2018models}. Results showed that Wikidata is unevenly distributed at an ontological level, with some domains having a lot of sub-classes automatically created by bots. 

\subsection{Type prediction in knowledge graphs}
Regarding works on type prediction tasks, SDType was one of the first tools made to add missing type statements in knowledge graphs \cite{paulheim2014improving} --specifically, in DBPedia. SDType works by calculating the statistical distribution of types in the object and subject position of a property and using this data to fill missing types of an entity, outperforming other type prediction approaches at the time. Melo et al. then proposed a new system that exploited the hierarchical nature of type systems in knowledge graphs by using a hierarchical classifier \cite{melo2017type}. This system was tested on several knowledge graphs, including Wikidata, outperforming SDType and other alternatives while also scaling better than other multilabel classifiers.

The use of knowledge graph embeddings for the task of type prediction has been explored recently. Kejriwal et al. proposed a method to create class embeddings based on a series of pre-computed entity embeddings \cite{kejriwal2017supervised}. Class embeddings are closer to related instances and similar classes in the vector space. Those authors then proposed the use of cosine similarity between the embedding of an entity and the class embeddings to recommend the typing of the entity. Other approaches try to generate fine-grained type predictions using entity embeddings in both supervised and unsupervised approaches, obtaining better results than SDType in two DBPedia datasets \cite{sofronova2020entity}. 

Ridle is a system that leverages neural networks and a custom knowledge graph representation to improve existing type prediction approaches \cite{weller2021predicting}. The proposed representation of each entity is built from its relations, under the idea that some sets of properties are more likely to appear as relations in the instances of a class than in others. Ridle outperforms SDType and other embedding-based approaches across multiple datasets, including Wikidata.

\subsection{Negative sampling methods}
Negative sampling methods have been shown to be an important step in the performance of knowledge graph embedding models. Kotnis et al. proposed two new embedding based negative sampling methods, comparing their performance to random approaches \cite{kotnis2017analysis}. Their results indicated that the performance of each negative sampling method depended on the characteristics of the dataset being used, with no approach being the best across all datasets.

Using generative adversarial networks (GANs) to produce negative samplers has also been introduced recently, showing promising results by improving the performance of knowledge graph embeddings under different settings \cite{cai2017kbgan, wang2018incorporating}. NSCaching is a method that improves the efficiency of GAN based methods by incorporating a cache of high-quality negative triples, outperforming state-of-the-art negative sampling methods \cite{zhang2019nscaching, zhang2021simple}.

\subsection{Leveraging the edit history of knowledge graphs}
Exploiting edit information of knowledge graphs for several tasks has been increasingly studied recently. Initial work tried to offer better ways to access and query edits from Wikidata. The Wikidata edit history query service was one of the first available tools, providing a SPARQL endpoint to query edit information made to Wikidata until July 1st of 2018 \cite{pellissier2019querying}. During the time of experimentation of this paper (November 2021 to March 2022) that service was down, but at the time of this writing it is back online.

Wikidated is a dataset that contains the entire edit history of Wikidata encoded as a set of triples additions and deletions \cite{schmelzeisen2021wikidated}. It provides a Python API for exploring and interacting with the dataset without needing to know the underlying data model being used. This dataset could not be used in our research since it is still undergoing active development.

Regarding the application of edit history information to knowledge graph refinement tasks, AlGhamdi et al. proposed a Wikidata recommendation system that leverages this information \cite{alghamdi2021learning}. Their system summarizes the editing activities of the user and combines this with entity representations to recommend items that the user could edit next. Results are promising and lay the ground for future work in Wikidata recommendation systems.

Finally, Tanon et al. proposed a system that automatically mines knowledge graph correction rules from its edit history \cite{pellissier2019learning}. Their system starts from a set of constraints violations from the knowledge graph, and mines the edit history to find edits that solved those violations in the past, which are then returned as a set of possible correction rules for the violation. They run a set of experiments in Wikidata, showing significant improvements over baselines. In their follow up work, they  proposed a refinement of their previous system by using neural networks to predict the correction rules of a given constraint \cite{pellissier2021neural}. That neural network receives as input a vector representation of the constraint, the triples that have been violated, and facts about the entities. Results over the Wikidata dataset showed significant improvements over both the previous rule-based approach and the baselines.

\section{Conclusions and future work}
\label{sec:conclusions}
This paper provides the first exploration of the use of edit history information in knowledge graph refinement and, specifically, type prediction tasks. We have built and provided a JSON dataset containing the edit history information of every Wikidata instance from the top 100 most important Wikidata classes based on their ClassRank score. This edit history information has been explored, making a special emphasis on its potential influence in knowledge graph refinement tasks. Finally, we have proposed two new approaches to improve knowledge graph embeddings in type prediction tasks: the use of negative samplers that generate corruptions based on the edit history of the graph and training a supervised model by labeling triples based on the edit history information. These approaches have been evaluated on a RDF dataset against currently used approaches in the field.

The main findings of our work are:
\begin{itemize}
    \item Instances in Wikidata have different edit behaviors based on the class they belong to. Some classes lead to more controversial statements that are the subject of edit wars, while some other classes are more stable and experience almost no removals or value replacements.
    \item Using removed triples from the graph as negative samples can improve the performance of a basic negative sampling approach in some cases, but in general it results in a worse performance of the models. On average, a basic negative sampling approach leads to a 123\% higher hits@1 score across every model compared to performing negative sampling from removed triples of the graph. Avoiding removed triples that experienced edit wars improves the performance of our proposed edit history based negative sampling approach across every knowledge graph embeddings model. On average, avoiding these triples improves the hits@5 of our proposed approach by 27.18\%.
    \item On the contrary, avoiding the generation of corruptions that belong to the set of removed triples from the graph improves the performance of 2 out of the 3 models, getting the 2 best results across every model and negative sampler combination being evaluated. This approach improves the hits@1 and hits@5 of the basic negative sampler approach by a 89.53\% and a 27.33\% respectively.
\end{itemize}

New promising research lines have emerged from this work, among them:
\begin{itemize}
    \item Reaching a deeper understanding on the differences between edits related to vandalism, a conflict of opinion within the community, or the natural evolution of an entity. Having a method to distinguish between such kinds of edits could be crucial to have a better understanding of the editing behavior in Wikidata and also to improve and create new methods that use this edit information.
    \item To perform an analysis of the corruptions generated by the negative samplers proposed in this work. This could provide a better understanding of the strengths and weaknesses of each model, while also providing an explanation of which types of edits should be avoided or considered by the negative samplers.
    \item There are also several optimizations that could be applied to the negative samplers proposed in this work. One of them could be to assign a score to each one of the corruptions generated by the negative samplers, and choosing the top $n$ corruptions based on this score instead of choosing them randomly.
    \item Performing an incremental evaluation of the performance of each negative sampler approach depending on the size of the dataset, and the number of negative triples from the edit history avoided or added. This evaluation could also include a comparison to more advanced negative sampling approaches (e.g., GAN-based negative samplers).
    \item Finally, the use of edit history information in other knowledge graph refinement tasks and models could also be explored. For example, one interesting research direction could be the creation of models that work directly with the dynamic version of the RDF dataset provided in this work.
\end{itemize}

\section*{CRediT author statement}
\textbf{Alejandro Gonzalez-Hevia:} Methodology, Conceptualization, Software, Validation, Formal Analysis, Investigation, Data Curation, Writing - Original Draft, Visualization, Project Administration. \textbf{Daniel Gayo-Avello:} Methodology, Resources, Writing - Review \& Editing, Supervision.

\section*{Declaration of Competing Interests}
The authors declare that they have no known competing financial interests or personal relationships that could have appeared to influence the work reported in this paper.

\bibliography{biblio}

\end{document}